\documentclass[11pt]{article}

\usepackage{acl}
\usepackage{times}
\usepackage{latexsym}
\usepackage[T1]{fontenc}
\usepackage[utf8]{inputenc}
\usepackage{microtype}
\usepackage{inconsolata}
\usepackage{graphicx}
\usepackage{booktabs}
\usepackage{multirow}
\usepackage{makecell}
\usepackage{amsmath,amssymb}
\usepackage{array}
\usepackage{xcolor}
\usepackage{tikz}
\usepackage{url}
\usepackage{xspace}
\usepackage{enumitem}
\usetikzlibrary{positioning}

\newcommand{\method}{\textsc{FWR}\xspace}
\newcommand{\longmean}{\textsc{LongMean}\xspace}

\definecolor{bluebox}{RGB}{231,240,251}
\definecolor{greenbox}{RGB}{232,246,236}
\definecolor{orangebox}{RGB}{253,242,224}
\usepackage{float}

\title{Write What Matters: Disentangling Token Selection from Slot Allocation in Recurrent Memory}

\author{
  Chen Chen \\
  Hong Kong Baptist University
  \And
  Rui Cheng \\
  Anhui University
}

\begin{document}
\maketitle

\begin{abstract}
% Long-sequence models must retain earlier information. Full attention keeps all past tokens but grows expensive, recurrent-memory models compress history into fixed slots for linear-time processing. Limited capacity makes writes consequential: distractors can interfere with useful content, while skipped evidence cannot be recalled. Existing multi-slot memories write every token to the same number of slots, merging whether to store it with how many slots to update. We introduce Factorized Write Routing (\method), which first keeps or skips each token and then chooses a fixed or token-adaptive number of slots; queries still read all slots.
% Across ten seeds on two synthetic tasks that query facts after long contexts, we observe a density-dependent reversal. With four slots per write ($K{=}4$), long-context accuracy favors always writing on dense Feature Recall (0.472 versus 0.337), but learned selective writing on sparse-fact Passkey (0.222 versus 0.048). The tested adaptive rule underperforms fixed allocation under the learned gate on both tasks. A diagnostic oracle given fact labels unavailable at deployment reaches 0.372 on Passkey, showing headroom in filtering distractors. The results suggest avoiding missed writes in information-dense contexts and filtering distractors when they dominate.
Long-sequence models must retain earlier information. Full attention keeps all past tokens but grows expensive, and recurrent-memory models compress history into fixed slots for linear-time processing. Limited capacity makes writes consequential: distractors can interfere with useful content, while skipped evidence cannot be recalled. Existing multi-slot memories write every token to the same number of slots, merging whether to store it with how many slots to update. We introduce Factorized Write Routing (FWR), which first keeps or skips each token and then chooses a fixed or token-adaptive number of slots. Queries still read all slots.
Across ten seeds on two synthetic tasks that query facts after long contexts, we observe a density-dependent reversal. With four slots per write, long-context accuracy favors always writing on dense Feature Recall (0.472 vs 0.337), but learned selective writing on sparse-fact Passkey (0.222 vs 0.048). The tested adaptive rule underperforms fixed allocation under the learned gate on both tasks. A diagnostic oracle given fact labels unavailable at deployment reaches 0.372 on Passkey, showing headroom in filtering distractors. The results suggest avoiding missed writes in information-dense contexts and filtering distractors when they dominate.
\end{abstract}

\section{Introduction}

Transformer attention provides random access to past tokens but incurs quadratic cost in sequence length \citep{vaswani2017attention}.  Linear attention, state-space models, and recurrent memory architectures instead compress history into a bounded state \citep{katharopoulos2020transformers,schlag2021linear,gu2024mamba,dao2024transformers}.  This compression enables constant-memory decoding, yet it creates a harder problem: once an update overwrites or contaminates the state, the original token may no longer be recoverable.  Empirical and theoretical work has consequently identified retrieval and copying as persistent weaknesses of fixed-state models \citep{arora2024zoology,jelassi2024repeat,wen2024rnns}.
Multi-slot recurrent memories mitigate this bottleneck by maintaining several independently updated states.  Recent systems route each token to a small Top-$K$ subset of slots \citep{zhang2024gated,afzal2026raven}.  Top-$K$ is an effective answer to \emph{where} a token should be written, but the usual fixed-$K$ rule silently assumes that every token should be written and deserves the same update budget.  That assumption is mismatched to long-context retrieval.  A context may be information-dense, in which nearly every token can be queried later, or information-sparse, in which a few facts are surrounded by distractors.  The same routing sparsity need not suit both regimes.

This paper asks a narrower and, we argue, more fundamental question: \textbf{In a bounded multi-slot memory, how should the model separate \emph{whether to write} a token from \emph{how many slots to update}, when the future query is not yet known?}
We study this question through Factorized Write Routing (\method), which decomposes routing into a binary write variable $z_t$ and a conditional slot count $K_t$.  The decomposition organizes controls along two explicit axes: always-write versus learned or oracle token selection, and fixed versus adaptive slot allocation.  It therefore exposes the two routing choices instead of packaging them in an inseparable ``adaptive router.''

Our results challenge two tempting generalizations.  First, sparse writing is not universally beneficial: on high-density Feature Recall, dense $K{=}M{=}16$ writing significantly outperforms $K{=}2$ at long lengths across ten paired seeds.  Second, token-adaptive $K$ is not sufficient by itself.  Pre-specified paired contrasts in the ten-seed mechanism study show a task-dependent reversal: at the same fixed width $K=4$, learned gating is worse than always writing on Feature Recall but better on sparse-fact Passkey.  Under learned gating, fixed allocation is favored over the tested adaptive rule on both tasks.  At $K=4$, an oracle token-type gate also outperforms the learned gate on Passkey, revealing headroom in fact--distractor selection without assuming knowledge of the eventual query.  Holding $K$ constant makes the fixed-family comparisons controlled tests of the gate policy within the tested recurrence.

Our contributions are threefold:
\begin{itemize}[leftmargin=*,nosep]
    \item We formulate multi-slot writing as two decisions: token selection and slot allocation, and provide \method, an implementation-compatible framework for independently controlling them.
    \item We give a ten-seed counterexample to the idea that a small fixed $K$ is a robust default: $K{=}2$ harms long recall when most context tokens remain relevant.
    \item Through pre-specified matched-$K$ paired contrasts and token-type oracle controls, we document a task-dependent reversal between learned and always-write gates, and show that our adaptive-$K$ rule underperforms validation-selected fixed $K$ under learned gating on both tasks.
\end{itemize}

\begin{figure*}[t]
\centering
\includegraphics[width=\textwidth]{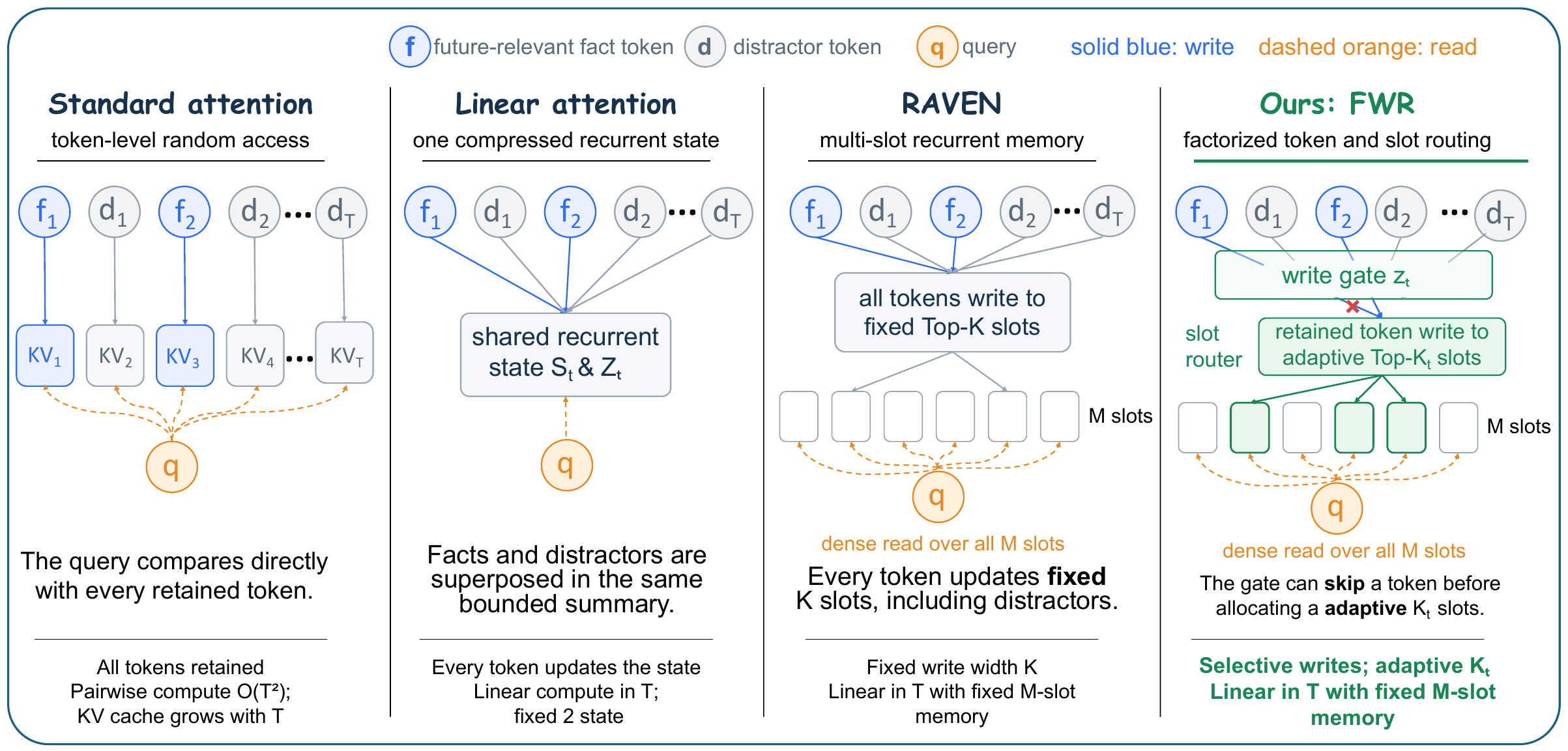}
\caption{Comparison of context storage and retrieval mechanisms between existing methods and ours. Standard attention retains token-level keys and values and allows the query to attend directly to all preceding tokens, whereas Linear attention compresses the context into a shared recurrent state to alleviate the huge KV cache memory. RAVEN maintains $M$ recurrent slots but routes every token to a fixed number $K$ of slots. Our Factorized Write Routing (FWR) first uses a token-level gate $z_t$ to decide whether to write, and then allocates the retained token to a token-adaptive number $K_t$ of slots.
Both RAVEN and FWR read densely over all $M$ slots, they differ in their write policies. The labels $f_i$ and $d_i$ illustrate tokens that are relevant and irrelevant, respectively, to the future query $q$; these relevance labels are unavailable to the learned gate at inference time.}
\label{fig:method-comparison}
\end{figure*}

\section{Related Work}

\paragraph{Efficient sequence models.}
Linear attention rewrites attention as a recurrent state update \citep{katharopoulos2020transformers} and can be interpreted as fast-weight programming \citep{schlag2021linear}.  Bounded-memory control improves the capacity and stability of this family \citep{peng2022abc}, while gated linear attention and delta-rule updates add data-dependent forgetting and error correction \citep{yang2024gla,yang2024delta,yang2024gateddelta}.  Selective state-space models similarly make recurrence input-dependent \citep{gu2024mamba}, and state-space duality connects their algorithms to structured attention \citep{dao2024transformers}.  Simple linear-attention recipes can achieve favorable recall--throughput trade-offs \citep{arora2024based}, but retrieval remains a central discriminator among efficient architectures \citep{arora2024zoology,jelassi2024repeat}.

\paragraph{Memory for long context.}
Transformer-XL recurs over cached hidden states \citep{dai2019transformerxl}; Compressive Transformer and Memorizing Transformer extend access through compressed or non-parametric memories \citep{rae2020compressive,wu2022memorizing}.  Recurrent Memory Transformer carries dedicated memory tokens across segments \citep{bulatov2022recurrent}, whereas Longformer uses sparse token-to-token attention \citep{beltagy2020longformer}.  These methods differ in how much past information remains directly addressable.  In bounded recurrent memory, writes must summarize or overwrite history, making routing quality as important as read quality.

\paragraph{Slot memories and conditional routing.}
Gated Slot Attention increases recurrent-state capacity with several independently updated slots \citep{zhang2024gated}.  RAVEN adds sparse routing and reports task-dependent optima for the number of updated slots \citep{afzal2026raven}.  Our work does not propose that dense or sparse routing universally supersedes RAVEN; it isolates a missing decision upstream of Top-$K$: whether a token should update memory at all.  Conditional computation offers related routing mechanisms.  Mixture-of-experts models send tokens to a small subset of parameters \citep{shazeer2017outrageously,fedus2022switch}, while adaptive computation and Mixture-of-Depths vary token-level computation \citep{graves2016adaptive,raposo2024mixture}.  Their objective is compute allocation; ours is the preservation of future-relevant information under a fixed recurrent state.

\paragraph{Long-context evaluation.}
LongBench and RULER demonstrate that nominal context length can overstate usable retrieval length \citep{bai2024longbench,hsieh2024ruler}, while position-controlled analyses expose failures when evidence lies away from prompt boundaries \citep{liu2024lost}.  We use synthetic tasks because they provide exact labels for future-relevant and distracting tokens, enabling direct measurement of write recall and false-positive rates.  This controlled diagnosis complements, but does not replace, evaluation on natural text.

\section{Factorized Write Routing}
\subsection{Method Overview}
\label{sec:method-overview}
Our method decomposes memory writing into two decisions: \emph{whether the current token should be stored} and \emph{how the write should be allocated across memory slots}.
This factorization allows the model to suppress uninformative tokens without constraining the slot-allocation mechanism used for informative ones.

Let $h_t\in\mathbb{R}^{d}$ denote the representation at sequence position $t$. The model maintains $M$ recurrent memory slots,
\begin{equation}
\mathcal{S}_{t-1}
=
\left\{
\mathbf{s}_{t-1}^{(m)}
\right\}_{m=1}^{M},
\end{equation}
where $\mathbf{s}_{t-1}^{(m)}$ is the content of slot $m$ before processing $h_t$. The number of slots $M$ is fixed, while their contents evolve as the sequence is processed.
For each token, a learned write gate first predicts
\begin{equation}
z_t\in\{0,1\},
\end{equation}
where $z_t=1$ indicates that the token should be written to memory and $z_t=0$ indicates that it should be skipped. In a separate routing branch, the slot router assigns scores to the memory slots:
\begin{equation}
a_{t,m}=\sigma(\ell_{t,m}),
\qquad m=1,\ldots,M,
\end{equation}
where $\ell_{t,m}$ is the routing logit for slot $m$.

The allocation module determines the number of slots $K_t$ to update and selects a corresponding slot set
\begin{equation}
\mathcal{I}_t\subseteq\{1,\ldots,M\},
\qquad |\mathcal{I}_t|=K_t.
\end{equation}
The allocation width $K_t$ may be fixed for all tokens or predicted adaptively from the routing scores. The selected scores are normalized to obtain the slot-allocation weights:
\begin{equation}
\bar r_{t,m}
=
\begin{cases}
\displaystyle
\frac{a_{t,m}}
{\sum_{j\in\mathcal{I}_t}a_{t,j}},
& m\in\mathcal{I}_t,\\[7pt]
0,
& m\notin\mathcal{I}_t.
\end{cases}
\label{eq:selected-route}
\end{equation}

The final write-routing weight combines token selection and slot allocation:
\begin{equation}
r_{t,m}=z_t\,\bar r_{t,m}.
\label{eq:route}
\end{equation}
Therefore, the routing weights satisfy
\begin{equation}
\sum_{m=1}^{M}r_{t,m}=z_t.
\end{equation}
When $z_t=0$, all routing weights are zero and the memory is carried forward unchanged. When $z_t=1$, the token is distributed over the selected slots according to their normalized routing scores.

Each slot is then updated through the recurrent memory operator:
\begin{equation}
\mathbf{s}_t^{(m)}
=
\mathcal{U}\!\left(
\mathbf{s}_{t-1}^{(m)},h_t,r_{t,m}
\right),
\qquad
\mathcal{U}(\mathbf{s},h,0)=\mathbf{s}.
\label{eq:update}
\end{equation}
Thus, only slots with nonzero routing weights are modified. Write routing determines where information is stored, whereas read attention independently determines which stored information is retrieved for the current query.

\begin{figure}[t]
\centering
\includegraphics[width=0.8\columnwidth]{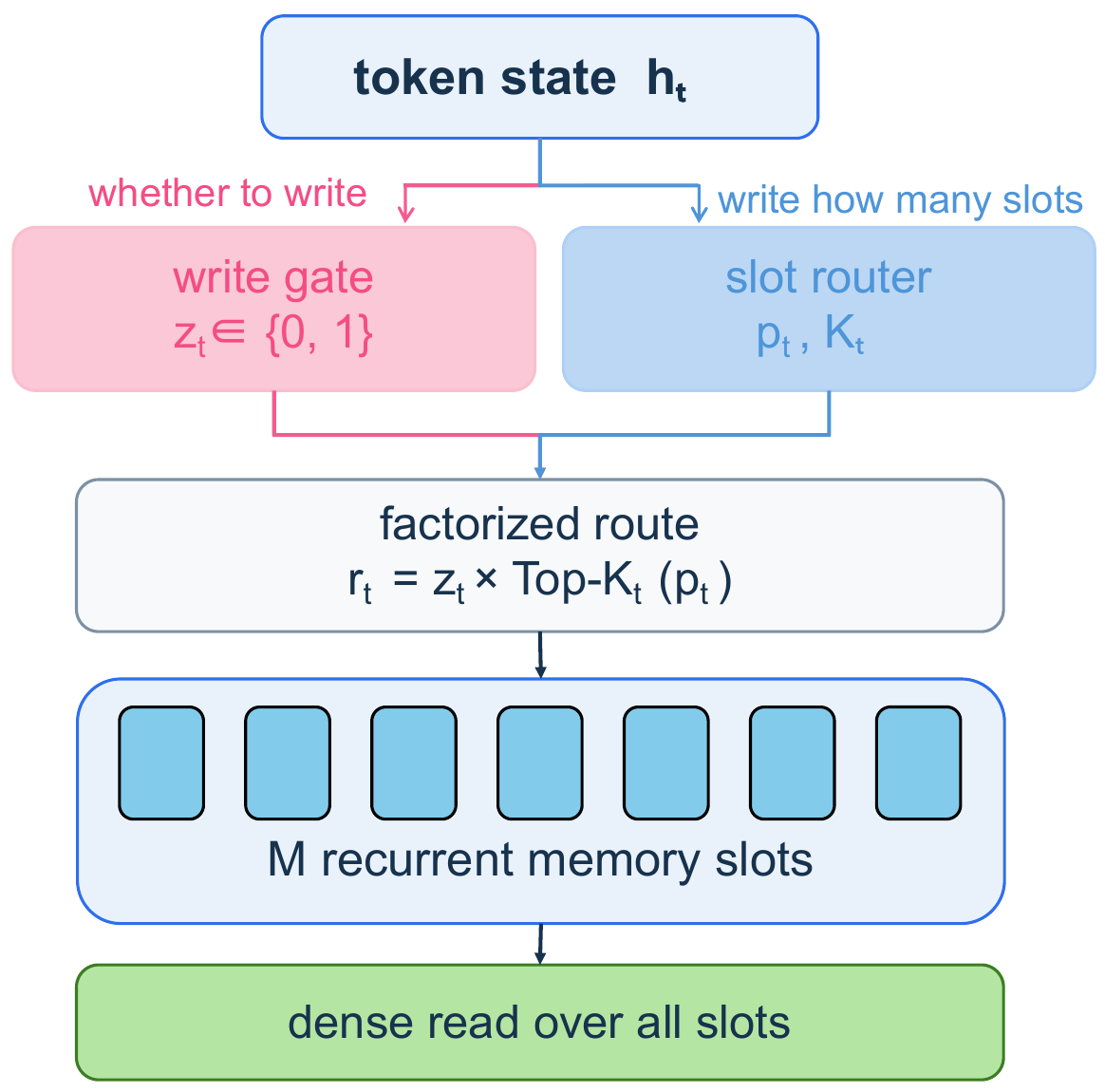}
\caption{\method separates token selection from slot allocation.  A skipped token changes no slot; a kept token updates either a fixed or token-adaptive number of slots.  Reading remains dense, so a query need not predict the write-time slot identity.}
\label{fig:method}
\end{figure}

\subsection{Token-Conditional Write Gate}
\label{sec:write-gate}

The write gate determines whether the current token should modify the memory. Given the token representation $h_t$, we compute its write probability as
\begin{equation}
q_t=\sigma\left(\mathbf{w}_z^\top h_t+b_z\right),
\label{eq:write-prob}
\end{equation}
where $\mathbf{w}_z$ and $b_z$ are learned parameters. The corresponding hard write decision is
\begin{equation}
z_t^{\mathrm{hard}}
=
\begin{cases}
1, & q_t\geq 0.5,\\
0, & q_t<0.5.
\end{cases}
\label{eq:hard-gate}
\end{equation}
Because the threshold operation is non-differentiable, we use a straight-through estimator during training:
\begin{equation}
z_t
=
q_t+
\operatorname{sg}\!\left(
z_t^{\mathrm{hard}}-q_t
\right),
\label{eq:ste-gate}
\end{equation}
where $\operatorname{sg}(\cdot)$ denotes the stop-gradient operator. Equation~\ref{eq:ste-gate} evaluates to $z_t^{\mathrm{hard}}$ in the forward pass, while its backward gradient is taken through $q_t$.

The gate is applied before slot allocation. If $z_t=0$, all routing weights are zero:
\begin{equation}
r_{t,m}=0,\qquad m=1,\ldots,M,
\end{equation}
and the memory remains unchanged. If $z_t=1$, the token is passed to the slot-allocation module, which determines how many slots should be updated and how the write should be distributed among them.

We encourage selective writing using the soft write-rate penalty
\begin{equation}
\mathcal{L}_{\mathrm{write}}
=
\frac{1}{T}\sum_{t=1}^{T}q_t.
\label{eq:write-regularization}
\end{equation}

\subsection{Fixed and Adaptive Slot Router}
\label{sec:adaptive-router}

The slot router first converts the positive routing scores $a_{t,m}=\sigma(\ell_{t,m})$ into a probability distribution over the $M$ memory slots:
\begin{equation}
\begin{aligned}
p_{t,m}
&= \frac{a_{t,m}}{\sum_{j=1}^{M} a_{t,j}} \\
&= \frac{\sigma(\ell_{t,m})}
        {\sum_{j=1}^{M} \sigma(\ell_{t,j})},
\quad m=1,\dots,M.
\end{aligned}
\label{eq:slot-probability}
\end{equation}
We write
$\mathbf p_t=(p_{t,1},\ldots,p_{t,M})$,
where $p_{t,m}\geq0$ and
$\sum_{m=1}^{M}p_{t,m}=1$.

For fixed-width routing, the number of selected slots is $K_t=K$. For adaptive routing, let
\begin{equation}
p_{t,(1)}
\geq p_{t,(2)}
\geq \cdots
\geq p_{t,(M)}
\end{equation}
denote the components of $\mathbf p_t$ sorted in descending order. We first find the smallest prefix whose cumulative probability reaches a mass threshold $\tau$:
\begin{equation}
\widehat K_t
=
\min\left\{
k\in\{1,\ldots,M\}:
\sum_{i=1}^{k}p_{t,(i)}
\geq\tau
\right\}.
\label{eq:mass-prefix}
\end{equation}
The actual allocation width is then bounded by
$K_{\min}$ and $K_{\max}$:
\begin{equation}
K_t
=
\min\left\{
K_{\max},
\max\left\{
K_{\min},\widehat K_t
\right\}
\right\}.
\label{eq:adaptive-k}
\end{equation}
The selected set $\mathcal I_t$ contains the $K_t$ slots with the largest routing probabilities. We use $\tau=0.8$, $K_{\min}=1$, and $K_{\max}=8$ in the adaptive-routing experiments. Because the hard selection of $K_t$ does not provide a convenient gradient for controlling allocation width, we use the entropy of the routing distribution:
\begin{equation}
H(\mathbf p_t)
=
-\sum_{m=1}^{M}
p_{t,m}\log p_{t,m}.
\label{eq:routing-entropy}
\end{equation}
We define the differentiable effective-slot count as
\begin{equation}
\widetilde K_t
=
\min\left\{
\exp\bigl(H(\mathbf p_t)\bigr),
K_{\max}
\right\}.
\label{eq:effective-k}
\end{equation}
The quantity $\widetilde K_t$ measures how many slots are effectively supported by the routing distribution.
Before clipping, \(\exp(H(\mathbf p_t))\) equals \(n\) for an ideal distribution that is uniform over \(n\) slots and zero elsewhere. We cap this value at \(K_{\max}\).
Unlike the discrete $K_t$, $\widetilde K_t$ is used only as a differentiable proxy in the slot-allocation regularizer $\mathcal L_K$.

\subsection{Training objective.}
Let $\mathcal Q$ denote the set of query positions in a mini-batch. For each $t\in\mathcal Q$, let $\mathbf{o}_t\in\mathbb{R}^{C}$ be the predicted answer logits and $y_t$ the target answer class. We train the model using the query-level cross-entropy loss
\begin{equation}
\mathcal{L}_{\mathrm{task}}
=
-\frac{1}{|\mathcal Q|}
\sum_{t\in\mathcal Q}
\log
\frac{\exp(o_{t,y_t})}
{\sum_{c=1}^{C}\exp(o_{t,c})}.
\label{eq:task-loss}
\end{equation}
This loss supervises only the answer predictions at query positions. The write gate and slot router receive no direct token-level relevance supervision. Instead, they are trained indirectly through the effect of earlier memory writes on subsequent query predictions.

We then define the normalized slot-allocation cost as
\begin{equation}
\mathcal L_K
=
\mathbb E_t
\left[
\frac{\widetilde K_t}{K_{\max}}
\right].
\label{eq:k-cost}
\end{equation}
Minimizing this term encourages the routing distribution to concentrate on fewer slots. In implementation, the expectation is additionally averaged over the examples and routing heads in each mini-batch.

The complete training objective is
\begin{equation}
\mathcal{L}
= 
\lambda_W\mathcal{L}_{\mathrm{write}}
+
\mathcal{L}_{\mathrm{task}}
+
\lambda_K\mathcal{L}_{K}
,
\label{eq:total-loss}
\end{equation}
where $\lambda_K\geq0$ and $\lambda_W\geq0$ control the penalties on slot-allocation width and write frequency, respectively.

\subsection{Dense Readout over Sparse Writes}
\label{sec:dense-read}

Although each token is written to only a subset of memory slots, a query can read from all $M$ slots. Let $u_t$ denote the query representation at position $t$. We compute a relevance score between the query and every slot:
\begin{equation}
e_{t,m}
=
u_t^\top
\phi_{\mathrm{key}}\!\left(\mathbf{s}_t^{(m)}\right),
\qquad m=1,\ldots,M,
\end{equation}
where $\phi_{\mathrm{key}}(\cdot)$ is a learned key projection. The read-attention weights are normalized over all slots:
\begin{equation}
\alpha_{t,m}
=
\frac{\exp(e_{t,m})}
{\sum_{j=1}^{M}\exp(e_{t,j})}.
\label{eq:read-attention}
\end{equation}
The retrieved representation is
\begin{equation}
c_t
=
\sum_{m=1}^{M}
\alpha_{t,m}
\phi_{\mathrm{value}}\!\left(\mathbf{s}_t^{(m)}\right),
\label{eq:read}
\end{equation}
where $\phi_{\mathrm{value}}(\cdot)$ is the corresponding value projection.

No predefined mapping assigns a fact or query to a particular slot. Instead, the write router, slot representations, and read attention are jointly optimized through the task loss. Consequently, a query does not need to know which slots were selected when a fact was written. It compares against all available slots at read time. The write width $K_t$ therefore controls how many slots are modified by the current token, and hence the potential write interference, but it does not restrict the $M$ slots available for subsequent retrieval.

\section{Experimental Setup}
\subsection{Tasks}
We construct key--value recall sequences followed by 16 queries.  Keys and values are sampled from vocabularies of size 1,024 and 64, respectively, and query tokens contain a key but mask its value.  Accuracy is exact value-token accuracy at query positions.
\paragraph{Write-gate controls.}
We compare the learned gate with two diagnostic controls. The always-write control sets $z_t=1$ for every token and therefore removes token-level filtering. The oracle control writes fact tokens and skips filler and query tokens using the ground-truth token-type labels provided by the synthetic data generator. The oracle is used only for diagnosis, it is unavailable in realistic inference and does not guarantee perfect accuracy because slot routing, memory updates, and retrieval may still fail.

\paragraph{Feature Recall (high density).}
Every context token is a unique key--value fact, and each query asks for a fact sampled from that context.  The model trains at context length 32 and is tested through 512.  Because nearly all context tokens are potential query targets, skipping writes risks false negatives.

\paragraph{Passkey (low density).}
Facts are placed randomly among filler tokens drawn from a disjoint key range.  Training uses 32 facts in 128 context tokens; testing scales both facts and context proportionally up to 256 facts in 1,024 tokens.  Only one quarter of context tokens are facts, so indiscriminate writing introduces avoidable memory interference.

Table~\ref{tab:tasks} summarizes the controlled difference.  Both tasks require length extrapolation, but they impose different future-relevance densities.

\begin{table}[t]
\centering
\small
\setlength{\tabcolsep}{3.5pt}
\begin{tabular}{lcc}
\toprule
 & Feature Recall & Passkey \\
\midrule
Training context & 32 & 128 \\
Training facts & 32 & 32 \\
Test contexts & 32--512 & 128--1,024 \\
Queries & 16 & 16 \\
Fact density & high & 25\% of context \\
Distractor tokens & none & 75\% of context \\
\bottomrule
\end{tabular}
\caption{The tasks share the same key--value recall objective but differ in future-relevant information density.}
\label{tab:tasks}
\end{table}

\subsection{Compared Methods and Evaluation}
\label{sec:comparison-setup}

All comparisons use the same RAVEN recurrent-memory backbone \citep{afzal2026raven}; the methods differ only in token selection and slot allocation. \textsc{Raven-Fixed} denotes the original always-write policy with a fixed Top-$K$ allocation, and \textsc{Raven-Dense} is its all-slot control with $K=M$. \textsc{FWR-NoGate} uses adaptive $K_t$ but
always writes, thereby isolating slot-count adaptation. \textsc{FWR-Fixed} and \textsc{FWR-Adaptive} use the learned write gate with fixed and adaptive allocation, respectively. The oracle variants replace the learned gate with ground-truth fact labels and are diagnostic, not deployable.

We conduct two complementary ten-seed comparisons. Table~\ref{tab:baseline} uses the official RAVEN backend and paired random seeds 1--10 to compare \textsc{Raven-Fixed} ($K=2$) with \textsc{Raven-Dense} ($K=M=16$). This experiment tests whether a small fixed $K$ is a reliable default; it does not use the proposed learned gate. Table~\ref{tab:factorial} uses the same seed set and a RAVEN-compatible recurrence to cross always-write, learned, and oracle gates with fixed or adaptive allocation. Hyperparameters are selected on validation data and fixed before test evaluation.

The primary metric is exact value-token accuracy at query positions. \longmean averages accuracy over context lengths strictly longer than training: 64--512 for Feature Recall and 256--1,024 for Passkey. Table~\ref{tab:factorial} reports the mean and sample standard deviation across ten seeds. For Table~\ref{tab:baseline} and the five pre-specified mechanism contrasts in Appendix Table~\ref{tab:paired-contrasts}, we compute differences within matched seeds and report paired 95\% Student-$t$ intervals and Cohen's $d_z$; ``favored'' means that the interval excludes zero. The three fixed-family contrasts hold $K=4$ constant and therefore compare gate policies directly within the tested recurrence. The adaptive-versus-fixed contrasts also change the realized write budget. Active writes is $\mathbb{E}[z_tK_t]$, and write fraction is $\mathbb{E}[z_t]$. Further implementation details appear in Appendix~\ref{app:implementation}.

\section{Results}
\label{sec:results}

\subsection{Fixed Top-\texorpdfstring{$K$}{K} versus All-Slot Writing}
\label{sec:fixed-vs-dense}

Table~\ref{tab:baseline} isolates slot-allocation width within the official
RAVEN backbone. \textsc{Raven-Fixed} writes every token to its top two
slots, whereas \textsc{Raven-Dense} writes every token to all $M=16$
slots. The two method columns report exact query accuracy at each context
length. We define
$\Delta=\textsc{Raven-Dense}-\textsc{Raven-Fixed}$, so positive values
favor all-slot writing. The paired confidence interval is computed from
the ten within-seed differences, and \longmean excludes the training
length.

\begin{table*}[t]
\centering
\caption{Ten-seed comparison of two RAVEN write-allocation rules.
Positive $\Delta$ favors \textsc{Raven-Dense}; ``dense favored'' means
that the paired 95\% confidence interval excludes zero. \longmean averages
only context lengths longer than training.}
\label{tab:baseline}
\small
\setlength{\tabcolsep}{4.0pt}
\begin{tabular}{llrrrrl}
\toprule
Task & Context length
& \shortstack{\textsc{Raven-Fixed}\\($K=2$)}
& \shortstack{\textsc{Raven-Dense}\\($K=M=16$)}
& $\Delta$
& \shortstack{Paired 95\% CI\\for $\Delta$}
& Evidence \\
\midrule
\multirow{6}{*}{Feature Recall}
& 32        & .7732 & .8800 & +.1068 & $[-.0536,.2673]$ & inconclusive \\
& 64        & .5663 & .7088 & +.1425 & $[-.0004,.2853]$ & inconclusive \\
& 128       & .3244 & .4426 & +.1182 & $[.0284,.2080]$  & dense favored \\
& 256       & .1698 & .2363 & +.0665 & $[.0212,.1118]$  & dense favored \\
& 512       & .0923 & .1212 & +.0289 & $[.0066,.0512]$  & dense favored \\
& \longmean & .2882 & .3772 & +.0890 & $[.0148,.1633]$  & dense favored \\
\midrule
\multirow{5}{*}{Passkey}
& 128       & .4267 & .3904 & -.0363 & $[-.2769,.2044]$ & inconclusive \\
& 256       & .2602 & .2793 & +.0191 & $[-.1576,.1958]$ & inconclusive \\
& 512       & .1392 & .1640 & +.0248 & $[-.0754,.1250]$ & inconclusive \\
& 1,024     & .0745 & .0818 & +.0074 & $[-.0416,.0563]$ & inconclusive \\
& \longmean & .1580 & .1751 & +.0171 & $[-.0900,.1242]$ & inconclusive \\
\bottomrule
\end{tabular}
\end{table*}

On high-density Feature Recall, \textsc{Raven-Dense} is favored at lengths 128, 256, and 512. Its \longmean is 0.3772, compared with 0.2882 for \textsc{Raven-Fixed}, giving a paired improvement of 0.0890, a 95\% CI of $[0.0148,0.1633]$, and Cohen's $d_z=0.86$. The intervals at lengths 32 and 64 include zero and are therefore inconclusive. Thus, writing every
token to only two slots can under-allocate memory capacity when nearly every context item may be queried later.

On Passkey, every confidence interval includes zero. \textsc{Raven-Dense} has a slightly higher \longmean than \textsc{Raven-Fixed} (0.1751 versus 0.1580), but their difference is only
0.0171 with a wide 95\% CI of $[-0.0900,0.1242]$. This experiment establishes neither a dense-writing advantage nor equivalence between the methods. Together, the two tasks reject the stronger claim that a small fixed $K$ is universally preferable.

\subsection{Separating Token Selection from Slot Allocation}
\label{sec:factorial-results}

Table~\ref{tab:factorial} compares the RAVEN controls with the proposed FWR variants. The Method, Gate, and Allocation columns state the routing factors used by each row. \longmean is long-context accuracy, active writes is the mean number of updated slots per token, and write fraction is the proportion of token--head decisions that trigger a write. Deterministic routing counts are shown without error bars.

\begin{table*}[t]
\centering
\caption{Ten-seed mechanism comparison on \longmean.
Non-deterministic entries report mean $\pm$ standard deviation
across matched seeds 1--10. Hyperparameters are fixed before
test evaluation. Bold marks the highest deployable mean per task.
Oracle methods ($\dagger$) use ground-truth token-type labels
and are diagnostic. Paired inference for five pre-specified contrasts
appears in Appendix Table~\ref{tab:paired-contrasts}.}
\label{tab:factorial}
\footnotesize
\setlength{\tabcolsep}{3.0pt}
\begin{tabular}{lllcrrr}
\toprule
Task & Method & Gate & Allocation
& \shortstack{\longmean$\uparrow$\\mean $\pm$ SD}
& \shortstack{Active writes\\mean $\pm$ SD}
& \shortstack{Write fraction\\mean $\pm$ SD} \\
\midrule

\multirow{7}{*}{Feature Recall}
& \textsc{Raven-Dense}
& Always & Dense ($K=16$)
& $\mathrm{.3855}\pm\mathrm{0.0124}$ & 16.000 & 1.000 \\

& \textsc{Raven-Fixed}
& Always & Fixed ($K=4$)
& $\mathrm{\textbf{.4716}}\pm\mathrm{0.0101}$ & 4.000 & 1.000 \\

& \textsc{FWR-NoGate}
& Always & Adaptive
& $\mathrm{.3978}\pm\mathrm{0.0150}$
& $\mathrm{5.488}\pm\mathrm{0.0182}$ & 1.000 \\

& \textsc{FWR-Fixed}
& Learned & Fixed ($K=4$)
& $\mathrm{.3369}\pm\mathrm{0.0216}$
& $\mathrm{1.602}\pm\mathrm{0.0872}$
& $\mathrm{.401}\pm\mathrm{0.0218}$ \\

& \textsc{FWR-Adaptive}
& Learned & Adaptive
& $\mathrm{.2776}\pm\mathrm{0.0206}$
& $\mathrm{.588}\pm\mathrm{0.0231}$
& $\mathrm{.179}\pm\mathrm{0.0184}$ \\

& \textsc{Oracle-Fixed}$^\dagger$
& Oracle & Fixed ($K=4$)
& $\mathrm{.3360}\pm\mathrm{0.0195}$
& $\mathrm{3.600}\pm\mathrm{0.0332}$
& $\mathrm{.900}\pm\mathrm{0.0125}$ \\

& \textsc{Oracle-Adaptive}$^\dagger$
& Oracle & Adaptive
& $\mathrm{.4217}\pm\mathrm{0.0264}$
& $\mathrm{4.294}\pm\mathrm{0.0251}$
& $\mathrm{.900}\pm\mathrm{0.0207}$ \\

\midrule

\multirow{7}{*}{Passkey}
& \textsc{Raven-Dense}
& Always & Dense ($K=8$)
& $\mathrm{.0456}\pm\mathrm{0.0024}$ & 8.000 & 1.000 \\

& \textsc{Raven-Fixed}
& Always & Fixed ($K=4$)
& $\mathrm{.0475}\pm\mathrm{0.0021}$ & 4.000 & 1.000 \\

& \textsc{FWR-NoGate}
& Always & Adaptive
& $\mathrm{.0609}\pm\mathrm{0.0026}$
& $\mathrm{4.493}\pm\mathrm{0.0278}$ & 1.000 \\

& \textsc{FWR-Fixed}
& Learned & Fixed ($K=4$)
& $\mathrm{\textbf{.2216}}\pm\mathrm{0.0082}$
& $\mathrm{.882}\pm\mathrm{0.0044}$
& $\mathrm{.221}\pm\mathrm{0.0035}$ \\

& \textsc{FWR-Adaptive}
& Learned & Adaptive
& $\mathrm{.1991}\pm\mathrm{0.0067}$
& $\mathrm{1.199}\pm\mathrm{0.0138}$
& $\mathrm{.299}\pm\mathrm{0.0046}$ \\

& \textsc{Oracle-Fixed}$^\dagger$
& Oracle & Fixed ($K=4$)
& $\mathrm{.3723}\pm\mathrm{0.0075}$
& $\mathrm{.965}\pm\mathrm{0.0124}$
& $\mathrm{.241}\pm\mathrm{0.0043}$ \\

& \textsc{Oracle-Adaptive}$^\dagger$
& Oracle & Adaptive
& $\mathrm{.3311}\pm\mathrm{0.0037}$
& $\mathrm{.849}\pm\mathrm{0.0052}$
& $\mathrm{.241}\pm\mathrm{0.0027}$ \\

\bottomrule
\end{tabular}
\end{table*}

\paragraph{High-density Feature Recall.}
Across ten seeds, the highest deployable mean is obtained by \textsc{Raven-Fixed} with $K=4$ ($0.4716\pm0.0101$), compared with $0.3855\pm0.0124$ for \textsc{Raven-Dense} and $0.3978\pm0.0150$ for \textsc{FWR-NoGate}. The learned-gate variants obtain $0.3369\pm0.0216$ and $0.2776\pm0.0206$, with write fractions of $0.401\pm0.0218$ and $0.179\pm0.0184$. At matched $K=4$, the pre-specified learned-fixed versus always-fixed contrast favors always writing ($\Delta=-0.1347$, CI $[-0.1481,-0.1213]$; Appendix Table~\ref{tab:paired-contrasts}), directly showing that the learned gate is harmful in this dense-relevance task. Together with Table~\ref{tab:baseline}, the results suggest that an intermediate fixed allocation can outperform either extreme in this setting, but do not establish a universal optimal $K$.

\paragraph{Low-density Passkey.}
The always-write variants remain low ($0.0456$--$0.0609$), whereas the learned-gate variants reach $0.2216\pm0.0082$ and $0.1991\pm0.0067$ while using $0.882\pm0.0044$ and $1.199\pm0.0138$ active writes. At matched $K=4$, the paired learned-fixed versus always-fixed contrast favors learned gating ($\Delta=0.1741$, CI $[0.1688,0.1794]$), isolating the benefit of selective writing in this sparse-fact task. Also at $K=4$, oracle-fixed exceeds learned-fixed ($\Delta=0.1507$, CI $[0.1451,0.1563]$). The oracle knows fact/filler type rather than the eventual query, so this gate-policy gap indicates token-selection headroom but does not measure future-query prediction.

\paragraph{Adaptive allocation is not consistently superior.}
The two pre-specified learned-gate allocation contrasts both favor fixed allocation: adaptive minus fixed is $-0.0593$ on Feature Recall (CI $[-0.0744,-0.0442]$) and $-0.0225$ on Passkey (CI $[-0.0279,-0.0171]$). Other gate conditions show mixed mean directions, so the adaptive rule has no task-independent advantage. Because the compared policies use different write budgets, these results reject a universal benefit for this adaptive-$K_t$ rule but do not prove that fixed allocation is intrinsically superior.

\section{Discussion}
\paragraph{A conditional routing hypothesis.}
The matched-$K$ gate contrasts reverse direction across tasks: learned-fixed is worse than always-fixed on Feature Recall but better on Passkey. Holding $K=4$ constant rules out allocation width as the explanation and supports write recall when nearly every token can be queried, but write precision when facts are embedded among distractors. The two tasks still differ in more than density, so these results do not identify density as the sole causal variable; a controlled continuous density sweep remains necessary.

\paragraph{What the oracle gap establishes.}
At time $t$, the writer observes the current prefix but not an arbitrary later query.  Our oracle supplies only generator-provided fact/filler labels: it writes every fact and skips fillers and queries, but it does not know the eventual query target.  The paired Passkey contrast holds $K=4$ fixed, removing allocation width as a confound, but query handling still differs between gates.  The gap therefore establishes headroom in the tested gate policy, not the value of perfect future-utility prediction.  Query-conditioned gates, read-only-query controls, and an oracle defined by eventual query membership are needed for the stronger claim.

\paragraph{Write--read consistency.}
Sparse writing does not require a query to reproduce the writer's discrete slot choice.  All slots participate in the read softmax (Eq.~\ref{eq:read}); training only needs useful information to induce a query-compatible key in at least one slot.  The unresolved risk is earlier: if $z_t=0$, no read rule can recover the omitted information.  This asymmetry further motivates recall-oriented supervision for the write gate.

\paragraph{Logical sparsity is not physical speed.}
Our implementation constructs sparse logical routes but currently passes a dense routing tensor to the recurrent kernel.  Consequently, $C_{\mathrm{logical}}$ characterizes update sparsity, not realized acceleration.  Claims about efficiency require a sparse kernel and direct measurements of latency, throughput, peak memory, and energy.  Accuracy--cost Pareto conclusions in this paper are therefore architectural, not systems-level.

\paragraph{Query-time updates are a confound.}
The recurrence updates memory before reading the current token.  Dense and always-write policies therefore allow query tokens themselves to alter the state; the oracle skips them, and a learned gate may do either.  Although all compared variants use the same base recurrence, part of the oracle gap could reflect query handling rather than context filtering alone.  Read-before-write and read-only-query ablations are necessary follow-up experiments.

\section{Conclusion}

We separated token selection from slot allocation in bounded multi-slot recurrent memory. The first paired ten-seed study shows that dense writing beats $K=2$ on high-density long recall. In the mechanism study, matched-$K$ ($K=4$) paired contrasts show that learned gating is worse than always writing on Feature Recall but better on sparse-fact Passkey; an oracle token-type gate further improves Passkey. Fixed allocation also beats the tested adaptive rule under learned gating on both tasks, although these allocation contrasts do not match realized write budgets. Overall, the evidence supports a conditional hypothesis---balance write precision and recall according to relevance density, and treat adaptive slot count as a separate capacity problem---that should next be tested with controlled density sweeps and natural text.

\section*{Limitations}

Both studies use ten paired seeds, and Appendix Table~\ref{tab:paired-contrasts} reports intervals for five pre-specified mechanism contrasts; it does not provide an exhaustive significance map for all rows in Table~\ref{tab:factorial}. The fixed-family contrasts hold $K=4$ constant, but the adaptive-versus-fixed contrasts use unequal write budgets, limiting causal attribution about allocation. The two synthetic tasks also differ in more than relevance density and omit ambiguous relevance, discourse, and natural-text generation. The oracle uses unavailable fact/filler labels, neither knows the future query nor constitutes an upper bound, and update-before-read confounds query handling. Scaling may change the trade-offs, while dense routing tensors make logical cost a sparsity proxy rather than hardware efficiency. Independent replication, equal-budget allocation controls, read-only-query ablations, and a continuous density sweep are needed.

\section*{Ethical Considerations}

The experiments use procedurally generated symbolic data and do not involve people, private information, or annotator labor.  The main downstream risk is overgeneralization: an oracle gate that recognizes task-defined facts should not be interpreted as evidence that demographic or clinical attributes are universally irrelevant.  In real applications, relevance depends on the user query, decision context, and fairness requirements.  Selective memory systems can silently discard information, so deployment in high-stakes domains would require domain-specific recall audits, subgroup analysis, traceable write decisions, and a fallback that preserves the original record.

\bibliography{references}

\appendix

\section{Reproducibility Details}

\paragraph{Data generation.}
Feature Recall samples unique context keys and random values; its 16 query keys are sampled from the context.  Passkey samples random filler keys, inserts fact key--value pairs at uniformly sampled context positions, and draws queries only from fact keys.  Fact and filler key ranges are disjoint for diagnosis.  Targets are defined only on the query suffix.

\paragraph{Length grids.}
Feature Recall trains and validates at length 32 and tests at 32, 64, 128, 256, and 512.  Passkey trains and validates with 32 facts in 128 context tokens and tests $(\text{context},\text{facts})\in\{(128,32),(256,64),(512,128),(1024,256)\}$.  All examples append 16 query tokens; reported context lengths exclude this suffix.

\paragraph{Randomness and hardware.}
Both the primary comparison and the mechanism study use seeds 1--10 with deterministic PyTorch execution on an NVIDIA RTX A6000 and the same optimizer schedule.  The captured environment reports PyTorch 2.13.0+cu130 and Flash Linear Attention 0.5.2.  Runs use bfloat16 and Gumbel-perturbed routing during training.

\paragraph{Implementation checks.}
Automated router mechanism tests pass.  With fixed $K$, the modified backend matches the official implementation exactly at sequence lengths 32 and 96 (maximum absolute and relative error zero).  For $M=8$, where the chunked kernel's backward path is unsupported in the tested library version, training and evaluation use the mathematically equivalent fused recurrent fallback; gradients are finite at length 96.

\section{Complete Variant Definitions}

\begin{table}[h]
\centering
\small
\setlength{\tabcolsep}{4pt}
\begin{tabular}{lll}
\toprule
Variant & Token selection & Slot allocation \\
\midrule
Dense & always write & fixed $K=M$ \\
Always/fixed & always write & fixed $K$ \\
Always/adaptive & always write & Eq.~\ref{eq:adaptive-k} \\
Learned/fixed & learned $z_t$ & fixed $K$ \\
Learned/adaptive & learned $z_t$ & Eq.~\ref{eq:adaptive-k} \\
Oracle/fixed & fact labels & fixed $K$ \\
Oracle/adaptive & fact labels & Eq.~\ref{eq:adaptive-k} \\
\bottomrule
\end{tabular}
\caption{The symmetric factorial design distinguishes token selection from slot allocation.}
\end{table}

\section{Further Diagnostic Results}

In a retained seed-1 routing diagnostic, the observed 10th/50th/90th percentiles of $K_t$ among written tokens were 1.00/2.55/7.89 for learned-gate Feature Recall, 3.00/5.36/8.00 for always-write Feature Recall, 2.50/4.00/5.50 for learned-gate Passkey, and 3.50/4.50/6.00 for always-write Passkey.  These ranges verify that the adaptive router does not collapse to a constant allocation in that run.  Because the percentile diagnostic has not been aggregated over all ten seeds, it is mechanistic evidence only; the accuracy conclusions rely on Table~\ref{tab:factorial}.

\section{Planned Confirmatory Tests}

A submission-strength follow-up should (i) replicate the five paired contrasts in Table~\ref{tab:paired-contrasts} on an independent seed set, (ii) sweep future-relevant token density while controlling fact count, distance, and query count, (iii) compare fixed and adaptive allocation at matched write budgets and cross $K_{\mathrm{train}}$ with $K_{\mathrm{infer}}$, (iv) compare update-before-read, read-before-write, and read-only query handling, and (v) add natural-text retrieval benchmarks plus measured latency and memory.

\section{Implementation and Configuration Details}
\label{app:implementation}

\paragraph{Model and optimization.}
All runs use a two-layer recurrent model with hidden dimension 256, four heads, an expansion factor of four, no positional encoding, and approximately 1.91M parameters. Key, value, and token-type embeddings form the input, and value embeddings are visible only at fact positions. Models are trained for 2,000 steps with AdamW \citep{kingma2015adam}, batch size 16, learning rate $10^{-3}$, weight decay 0.01, gradient clipping at 1.0, and bfloat16 arithmetic. Gumbel perturbations \citep{jang2017categorical} are applied to routing logits during training. Checkpoints are selected by validation accuracy, with validation loss used to break ties.

\paragraph{Routing search space.}
The fixed-width grid is $K\in\{2,4,6,8\}$, subject to $K\leq M$; dense routing uses $K=M$. Adaptive routing uses $K_{\min}=1$, $K_{\max}=\min(8,M)$, mass threshold $\tau=0.8$, and gate threshold 0.5. For the learned-gate and no-gate adaptive variants, we search $\lambda_K\in\{0,0.001,0.01\}$. The reported mechanism study uses $\lambda_W=0$. Within each task and method family, we retain configurations within 0.005 validation accuracy of the family best and select the one with lowest $C_{\mathrm{logical}}$.

\begin{table}[t]
\centering
\caption{Validation-selected configurations for the mechanism study in Table~\ref{tab:factorial}. These are configuration details, not additional test results.}
\label{tab:selected-configurations}
\small
\setlength{\tabcolsep}{3.7pt}
\begin{tabular}{lll}
\toprule
Task & Method & Selected config. \\
\midrule
\multirow{7}{*}{Feature Recall}
& \textsc{Raven-Dense}      & $K=16$ \\
& \textsc{Raven-Fixed}      & $K=4$ \\
& \textsc{FWR-NoGate}       & $\lambda_K=.001$ \\
& \textsc{FWR-Fixed}        & $K=4$ \\
& \textsc{FWR-Adaptive}     & $\lambda_K=0$ \\
& \textsc{Oracle-Fixed}     & $K=4$ \\
& \textsc{Oracle-Adaptive}  & $\tau=.8$ \\
\midrule
\multirow{7}{*}{Passkey}
& \textsc{Raven-Dense}      & $K=8$ \\
& \textsc{Raven-Fixed}      & $K=4$ \\
& \textsc{FWR-NoGate}       & $\lambda_K=0$ \\
& \textsc{FWR-Fixed}        & $K=4$ \\
& \textsc{FWR-Adaptive}     & $\lambda_K=0$ \\
& \textsc{Oracle-Fixed}     & $K=4$ \\
& \textsc{Oracle-Adaptive}  & $\tau=.8$ \\
\bottomrule
\end{tabular}
\end{table}

\paragraph{Evaluation and implementation checks.}
The primary comparison uses $M=16$, seeds 1--10, and 30 evaluation batches per test condition. The mechanism study also uses seeds 1--10, with $M=16$ for Feature Recall, $M=8$ for Passkey, and 10 evaluation batches per condition. Runs use deterministic PyTorch execution on an NVIDIA RTX A6000. The modified fixed-$K$ path matches the official RAVEN implementation exactly at sequence lengths 32 and 96. For $M=8$, we use the mathematically equivalent fused recurrent kernel because the tested chunked-kernel backward path does not support that slot dimension. The two-sided intervals in Tables~\ref{tab:baseline} and~\ref{tab:paired-contrasts} are Student-$t$ intervals over ten within-seed differences; Table~\ref{tab:factorial} reports sample standard deviations across seeds.

\section{Auxiliary Comparison of Single-State and Multi-State Memory}
\label{app:crla-pilot}

\begin{table*}[!t]
\centering
\caption{Auxiliary comparison of single-state and multi-state memory under the earlier prototype protocol. The metric is best validation accuracy, not the \longmean test metric used in the main experiments. Three-run entries are mean $\pm$ standard deviation; rows without error bars contain one run. GLA-style is an internal lightweight gated baseline rather than an official GLA implementation. Bold marks the highest observed value within each pilot task, not a statistically established winner.}
\label{tab:crla-pilot}
\small
\setlength{\tabcolsep}{4pt}
\begin{tabular}{lllcr}
\toprule
Pilot task & Train--evaluation setting & Model and memory organization
& Runs & Best val.\ accuracy \\
\midrule
\multirow{3}{*}{Feature Recall}
& \multirow{3}{*}{32 $\rightarrow$ 256 facts}
& Linear: one state & 1 & .661 \\
& & GLA-style: one gated state & 3 & $.673\pm.005$ \\
& & CRLA: 16 soft-routed states & 3 & $\mathbf{.861\pm.005}$ \\
\midrule
\multirow{4}{*}{Feature-Passkey}
& \multirow{4}{*}{\shortstack{context/facts:\\
128/32 $\rightarrow$ 512/128}}
& Linear: one state & 1 & .678 \\
& & GLA-style: one gated state & 3 & $.715\pm.010$ \\
& & CRLA: 16 soft-routed states & 3 & $\mathbf{.818\pm.029}$ \\
& & Softmax: full-prefix access & 3 & $.805\pm.043$ \\
\bottomrule
\end{tabular}
\end{table*}

\begin{table*}[t]
\centering
\caption{Pre-specified paired contrasts across seeds 1--10 for the
validation-selected configurations in Table~\ref{tab:factorial}.
$\Delta>0$ favors the first method. Confidence intervals are
two-sided Student-$t$ intervals over seed-wise differences. The three
fixed-allocation contrasts hold $K=4$ constant.}
\label{tab:paired-contrasts}
\small
\setlength{\tabcolsep}{4pt}
\begin{tabular}{llrrrr}
\toprule
Task & Contrast
& $\Delta$\longmean
& Paired 95\% CI
& $d_z$
& Evidence \\
\midrule
Feature Recall
& Learned vs.\ always, fixed $K=4$
& $-0.1347$
& $[-0.1481,-0.1213]$
& $-7.20$
& Always favored \\

Feature Recall
& Adaptive vs.\ fixed, learned gate
& $-0.0593$
& $[-0.0744,-0.0442]$
& $-2.81$
& Fixed favored \\

Passkey
& Learned vs.\ always, fixed $K=4$
& $+0.1741$
& $[0.1688,0.1794]$
& $23.60$
& Learned favored \\

Passkey
& Adaptive vs.\ fixed, learned gate
& $-0.0225$
& $[-0.0279,-0.0171]$
& $-2.98$
& Fixed favored \\

Passkey
& Oracle vs.\ learned, fixed $K=4$
& $+0.1507$
& $[0.1451,0.1563]$
& $19.14$
& Oracle favored \\
\bottomrule
\end{tabular}
\end{table*}

The main experiments begin with a memory divided into $M$ recurrent slots. Within this setting, FWR asks two questions: whether the current token should write to memory, represented by $z_t$, and, conditional on writing, how many slots it should update, represented by $K_t$. This appendix steps back to a more basic architectural question: why use multiple recurrent memory slots in the first place? Specifically, we examine whether separating one recurrent state into several routed states can improve associative recall.

We study this question using an earlier auxiliary prototype, Capacity-Routed Linear Attention (CRLA). The Linear and GLA-style controls maintain one recurrent state per head. CRLA-16 instead maintains 16 such states and uses soft routing weights to distribute every token across them. In the notation of the main paper, CRLA has neither a binary write decision $z_t$ nor a discrete allocation width $K_t$: every token writes, and all 16 states receive weighted updates. CRLA also uses a different recurrent update from RAVEN. Thus, this
pilot compares one recurrent state with multiple softly routed states, whereas the main RAVEN/FWR experiments compare write policies after the $M$-slot memory architecture has been fixed.

The pilot predates the controlled RAVEN study and differs from it in model scale, recurrent update, checkpoint selection, seed coverage, and evaluation metric. We therefore report it separately and do not compare its numerical values directly with Tables~\ref{tab:baseline} and~\ref{tab:factorial}.

Within this separate protocol, the 16-state prototype has the highest observed validation accuracy on both pilot tasks. On Feature-Passkey, its observed accuracy is also numerically similar to softmax attention. This pattern is consistent with the motivation for using multiple recurrent memory slots: separating stored associations across routed states may reduce interference. It does not, however, establish a controlled architectural ranking. The seed coverage is unbalanced, the metric is selected on validation data, and CRLA differs from the controls in more than state count alone. Most importantly, the pilot does not test either the FWR write gate or adaptive-$K_t$ allocation; its role is only to motivate the multi-slot starting point used by the main RAVEN/FWR study.

\end{document}